\documentclass[10pt,twocolumn,letterpaper]{article}

\usepackage[pagenumbers]{wacv} 

\usepackage{algorithm}
\usepackage{algpseudocode} 
\usepackage{mathtools}
\usepackage{amsthm}
\usepackage{bm}
\usepackage{multirow}  
\usepackage[accsupp]{axessibility}

\newtheorem{theorem}{Theorem}  

\definecolor{wacvblue}{rgb}{0.21,0.49,0.74}
\usepackage[pagebackref,breaklinks,colorlinks,allcolors=wacvblue]{hyperref}

\def\wacvPaperID{1719} 
\def\confName{WACV}
\def\confYear{2026}

\title{FedEFC: Federated Learning Using Enhanced Forward Correction\\
Against Noisy Labels}

\author{Seunghun Yu\\
KAIST\\
Daejeon, South Korea\\
{\tt\small sh0703.yu@kaist.ac.kr}
\and
Jin-Hyun Ahn\\
Myongji University\\
Yongin, South Korea\\
{\tt\small wlsgus3396@mju.ac.kr}
\and
Joonhyuk Kang\\
KAIST\\
Daejeon, South Korea\\
{\tt\small jkang@kaist.ac.kr}
}

\begin{document}
\maketitle
\begin{abstract}
Federated Learning (FL) is a powerful framework for privacy-preserving distributed learning. 
It enables multiple clients to collaboratively train a global model without sharing raw data.
However, handling noisy labels in FL remains a major challenge due to heterogeneous data distributions and communication constraints, which can severely degrade model performance.
To address this issue, we propose \texttt{FedEFC}, a novel method designed to tackle the impact of noisy labels in FL. \texttt{FedEFC} mitigates this issue through two key techniques: (1) prestopping, which prevents overfitting to mislabeled data by dynamically halting training at an optimal point, and (2) loss correction, which adjusts model updates to account for label noise. 
In particular, we develop an effective loss correction tailored to the unique challenges of FL, including data heterogeneity and decentralized training. 
Furthermore, we provide a theoretical analysis, leveraging the composite proper loss property, to demonstrate that the FL objective function under noisy label distributions can be aligned with the clean label distribution.
Extensive experimental results validate the effectiveness of our approach, showing that it consistently outperforms existing FL techniques in mitigating the impact of noisy labels, particularly under heterogeneous data settings (e.g., achieving up to 41.64\% relative performance improvement over the existing loss correction method).
\end{abstract}    
\section{Introduction}
\label{sec:intro}

Federated Learning (FL) is a powerful paradigm for distributed learning that enables the training of a high-performing global model without requiring the aggregation or centralization of locally stored data \cite{mcmahan2017communication}. While FL provides strong privacy guarantees by keeping client data decentralized, this non-centralized nature makes the model highly sensitive to the underlying data distribution among clients. The challenge of convergence under heterogeneous data distributions has been extensively studied \cite{reddi2021convergence, li2021ditto, li2020fedprox, karimireddy2020scaffold, acar2020fedDyn}.

Recently, a growing body of research \cite{yang2022robust, Xu_2022_CVPR, fang2022robust, wu2023fednoronoiserobustfederatedlearning, Bixiao_2022_ACMT} has investigated the impact of noisy datasets in addition to data heterogeneity. 
The adverse effects of label noise—whether caused by natural annotation errors—are often more pronounced in FL than in centralized learning (CL) due to the decentralized nature of training and the aggregation of corrupted model updates. Moreover, the lack of direct access to client datasets significantly limits the applicability of conventional noise mitigation methods commonly used in CL, necessitating novel approaches tailored to the FL setting.

\begin{figure*}[ht]
    \centering
    \includegraphics{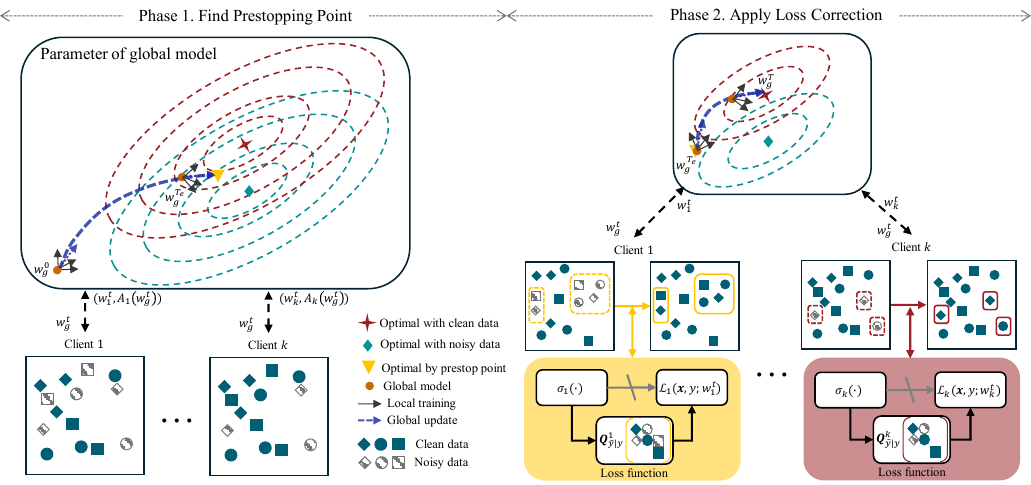} 
    \caption{
    Overview of \texttt{FedEFC} framework. 
    The scheme consists of two phases: (1) determining the prestopping point and (2) refining the loss correction.
    In Phase 1, the centralized server tracks client training accuracies to identify the prestopping point where model parameters are near-optimal. 
    In Phase 2, each client updates its loss function using enhanced forward correction, guiding global model parameters toward their optimal configuration in the clean data space.
    }
    \label{fig:proposed_diagram}
\end{figure*}

Building on this perspective, we propose \texttt{FedEFC}, an effective FL algorithm for mitigating the impact of noisy datasets. Our approach utilizes two key techniques:
\begin{itemize}
    \item \textit{Prestopping} : A dynamic early stopping mechanism that prevents overfitting to mislabeled data by halting training at an optimal point.
    \item \textit{Loss Correction} : A robust adjustment of model updates to account for label noise, ensuring mitigation of noisy labels after the prestopping point.
\end{itemize}
Here, the proposed loss correction method is applied after the prestopping phase, replacing the standard update. Notably, our loss correction technique is specifically designed for consistent effectiveness in heterogeneous FL settings by refining and extending the forward correction method proposed in \cite{Patrini_2017_CVPR}. To achieve this, we introduce an alternative estimation process that improves the accuracy of the noise transition matrix and dynamically updates the loss function, leading to an enhanced forward correction mechanism. The overall architecture of \texttt{FedEFC} is depicted in \cref{fig:proposed_diagram}. Our main contributions in this work are as follows:
\begin{itemize}
    \item We develop an enhanced forward correction to mitigate the impact of noisy labels without directly altering the data, thereby preventing unnecessary information loss (\cref{sec:main}). When integrated with the prestopping technique, this approach effectively reduces the adverse effects of noisy labels, particularly in heterogeneous FL settings.
    \item We provide a theoretical proof demonstrating that the enhanced forward correction enables each client to achieve the comparable training effectiveness as if learning from entirely clean data, despite noisy labels. (\cref{sec:main}).
    \item Experimental results confirm that \texttt{FedEFC} outperforms existing FL techniques, demonstrating its robustness in mitigating the impact of noisy labels and ensuring reliable model performance (\cref{sec:experiments}). To achieve this, we incorporate sparsity \cite{Curtis_2021_JAIR}, a measure that quantifies the degree of asymmetric label noise, into our noisy label synthesis process. Additionally, we leverage the Dirichlet distribution and Bernoulli distribution to systematically allocate data in a heterogeneous manner, ensuring a realistic simulation of non-IID conditions in FL. (\cref{sec:preliminaries}).
\end{itemize}

\subsection{Related Works}
\textbf{Confident learning}: 
As data-centric AI has gained prominence over model-centric approaches, effectively handling noisy labels has become increasingly critical, particularly when working with large-scale datasets \cite{russakovsky2015imagenet, lin2014microsoft}. Numerous studies have investigated techniques for identifying and managing mislabeled data \cite{vanrooyen2015learning, chen2019understanding, han2018coteaching, Curtis_2021_JAIR}.  Among these, \texttt{confident learning} \cite{Curtis_2021_JAIR} leverages a count matrix to model the relationship between true and noisy labels. This matrix has been demonstrated to be highly effective in refining mislabeled instances within noisy datasets. In this work, we incorporate the count matrix into \texttt{FedEFC} to further enhance the forward correction, improving robustness against label noise in heterogeneous setting. 

In \cite{Bixiao_2022_ACMT}, count matrix has also been integrated into FL for label correction. However, this approach has notable limitations, as it does not explicitly account for heterogeneous data distributions across clients and depends on a pretrained model for reliable performance, which may not always be feasible in real-world FL scenarios. Though, since adapting \texttt{confident learning} in FL offers a meaningful baseline for comparison with \texttt{FedEFC}, we modify it to ensure a fair and consistent evaluation within the FL setting.

\textbf{Forward loss correction}:
\texttt{forward correction} \cite{Patrini_2017_CVPR} is one of the main approaches designed to mitigate the detrimental effects of noisy labels by adjusting model predictions based on an estimated noise transition matrix. Both theoretical analysis and experimental results have validated the effectiveness of loss correction, demonstrating that it enables training on noisy datasets to approximate the learning dynamics of training on clean datasets. Nevertheless, a major challenge remains in constructing the reliable noise transition matrix, which is crucial for effective correction. In this work, \texttt{FedEFC} utilizes forward loss correction for the clients' local training of FL, while the estimation of noise transition matrix is tailored to be robust in the heterogeneous FL settings. To enhance reliability, the matrix is re-estimated at each training round after the prestopping point, leveraging the temporal global model. Furthermore, for estimation, we apply the manner of count matrix, instead of the model prediction.
As shown later, this method provides a more stable estimation, particularly in heterogeneous environments. A detailed discussion of this approach is provided in \cref{sec:main}.

\textbf{FL algorithms for non-IID settings}: In real-world FL scenarios, data is typically distributed in a non-IID manner, introducing significant challenges for model training \cite{hsu2019measuring, zhao2018federated}. Various strategies have been proposed to enhance FL performance under heterogeneous settings. 
For instance, \texttt{FedProx} \cite{li2020fedprox} introduces minor modifications to \texttt{FedAvg} to achieve more robust convergence when training on non-IID data.
\texttt{Ditto} \cite{li2021ditto} enhances data robustness and fairness, improving individual client performance while maintaining overall model consistency. In this work, we consider these methods as baseline algorithms for comparison with \texttt{FedEFC}, evaluating its effectiveness in mitigating noisy labels and handling heterogeneous FL settings.

\textbf{FL algorithms against noisy dataset}: Several research efforts have explored ways to resolve the noise labels present in local datasets of FL.
In \cite{Xu_2022_CVPR}, \texttt{FedCorr} is designed to cope with noisy data challenges.
On the other hand, the framework requires the assumption that certain clients have entirely \textit{clean} data to ensure improvement.
\texttt{FedNoRo} \cite{wu2023fednoronoiserobustfederatedlearning} handles class imbalance and heterogeneous label noise in FL, but it requires accurate \textit{identification} of noisy clients, which may be challenging in practice.
Although \texttt{RoFL} \cite{yang2022robust} effectively leverages similarity-based learning to mitigate the issue of asymmetric noisy labels, it requires additional information on \textit{feature} data that may pose a potential risk to privacy and communication bottleneck. 
All considered approaches, including \texttt{FedCorr}, are compared with the proposed method to provide a comprehensive and fair evaluation.

\section{Preliminaries}
\label{sec:preliminaries}
In this section, we describe the fundamentals of FL, including weight aggregation, data allocation, and noise generation, as applied in an FL scenario. 
In particular, we focus on non-IID data allocation and a practical noise generation method to reflect real-world settings.

In an FL environment, we assume a federated network with a centralized server and $N$ clients.
The clients are elements of the set $\mathcal{S}$, where $|\mathcal{S}| = N$. 
Each client $k$ is assigned a local dataset denoted as $\mathcal{D}_k = \{(\boldsymbol{x}^k, \tilde{y}^k)\}^{n_k}$, consisting of $n_k$ examples.  
Each example pairs an input sample $\boldsymbol{x}^k$ with its observed noisy label $\tilde{y}^k$.   
The union of all local datasets is expressed as $\mathcal{D} = \bigcup_{k=1}^{N} \mathcal{D}_k$, but each client's data remains private and is never shared with the server.
All labels in the dataset belong to the label set $\mathcal{C}$, where $\tilde{y}^{k} \in \mathcal{C}$.

During each local training round, we define the local model weights for client $k$ as $w_k^{t}$, where $t$ represents the training round. 
The central server aggregates the weights from the participating clients in each round. 
The set of participating clients in round $t$ is denoted as $\mathcal{S}_t$ where $\mathcal{S}_t \subseteq \mathcal{S}$. 
Therefore, the global model weights $w_g^t$ at round $t$ are aggregated as follows, as described in \cite{mcmahan2017communication}
\begin{equation}
w_g^t \leftarrow \sum_{k \in \mathcal{S}_t} \frac{n_k}{\sum_{u \in \mathcal{S}_t} n_u} w_{k}^{t}.
\end{equation}

\subsection{Data Allocation} 
We consider a non-IID data allocation strategy to reflect practical scenarios.
Even in a non-IID setting, certain configurations can approximate IID characteristics by varying the balancing parameters.
Two key parameters influence non-IID allocation: $\alpha_\text{dir}$ is derived from the Dirichlet distribution, which determines the number of examples assigned to each client, while $p$ is based on the Bernoulli distribution, which controls the label distribution among the examples allocated to clients.  
As described in \cite{Xu_2022_CVPR}, a combination of the Dirichlet and Bernoulli distributions is used to construct non-IID data distributions.

Specifically, the Bernoulli probability $p$ determines whether examples of a specific label $i$ are allocated to client $k$. 
The Bernoulli distribution is represented by the indicator $\mathbb{I}(k, i) \sim p$, which takes the value $1$ if label $i$ is allocated to client $k$, and $0$ otherwise. 
Once the Bernoulli distribution is established, the number of examples assigned to clients with $\mathbb{I}(\cdot, i) = 1$ follows a Dirichlet distribution parameterized by $\alpha_{\text{dir}} > 0$.  
Therefore, the degree of non-IID data allocation is controlled by the parameters $p$ and $\alpha_{\text{dir}}$. 

\subsection{Noise Generation}
A symmetric noise flip more closely resembles real-world scenarios than the symmetric case when generating synthetic noise.
To generate asymmetric noise, the sparsity mentioned in \cite{Curtis_2021_JAIR}, \cite{Tsouval_2024_ACMT} is utilized. 
The amount of noise $\rho$ and sparsity $\zeta$ need to be considered for noise generation. 
$\rho$ is the fraction of samples whose labels are flipped while $\zeta$ represents the proportion of uncorrupted labels, excluding the true label. 
For example, if all labels are flipped, $\zeta = 0$.
In contrast, $\zeta=1$ indicates that the dataset contains only clean labels.
Thus, high sparsity represents greater imbalance in noisy labels.

\section{Proposed Method}\label{sec:main}
In this section, we propose \texttt{FedEFC} for mitigating the impact of noisy labels in non-IID FL.
\texttt{FedEFC} consists of two phases: (1) determining a prestopping point and (2) applying loss correction.
In Phase 1, the centralized server estimates the accuracy of the global model by aggregating the training accuracies of local clients.
During Phase 1, each client transmits its measured accuracy to the centralized server, which monitors accuracy variations to determine the prestopping point. 
Once this point is reached, the server notifies the clients that Phase 1 is complete and that no further accuracy updates will be transmitted in subsequent rounds.
In Phase 2, participating clients generate a noise transition matrix by analyzing the allocated dataset and apply loss correction.
Further details are provided below.

\subsection{Finding the Prestopping Point}
In Phase 1, we explore the learning property when the dataset includes noisy labels.
As observed in \cite{song2020earlystopping}, the loss of a model tends to sharply increase after a certain epoch when trained on a dataset with noisy labels.
This phenomenon indicates that beyond prestopping point, the model can no longer effectively learn from clean data and instead begins to overfit to noisy labels \cite{Li_2020_AISTATS}.
As the accuracy of the global model fluctuates and saturates, clients experience a degraded learning performance on their local datasets.
However, heuristic validation from a client's perspective is difficult to adopt in FL.
Therefore, we introduce the heuristic validation method suitable for an FL system.

\begin{figure}[t] 
  \centering
  \includegraphics{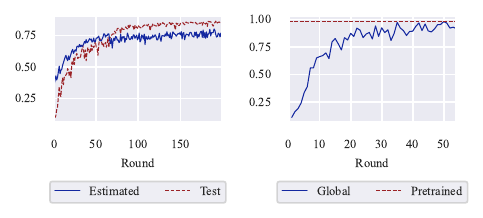}
  \caption{
  Left: Test accuracy and estimated accuracy used to determine the prestopping point.
  Right: Cosine similarity between the real noise matrix and the noise matrices estimated by the pretrained model and the global model in training.
  }
  \label{fig:fig1}
\end{figure}

We denote the estimated accuracy of the global model in round $t$ as follows
\begin{align}\label{eq:accuracy}
A(t) = \sum_{k\in\mathcal{S}_t} \frac{A_k(w_{g}^t)}{|\mathcal{S}_t|},
\end{align}
where $A_k(w_{g}^t)$ is training accuracy of client $k$ with the global model based on its own dataset.
The training set contains both clean and noisy labels, which differs from the original heuristic validation approach that relies on a clean set \cite{song2020earlystopping}.
A threshold $\gamma_{\text{thr}}$ is defined to quantify the instability in learning.
In addition, a patience parameter $\tau_{p}$ is employed to track previous accuracy history.
The accuracy measured in the current round, \(A(t)\), is compared to the highest observed accuracy \(A_\text{max}\), where $A_\text{max} = \max_{i < t} A(i)$ with $A_\text{max}$ initialized to $0$.
If \(A(t) < A_\text{max}\), the patience parameter \(\tau_p\) is incremented by 1; otherwise, \(\tau_p\) is reset to $0$ and \(A_\text{max}\) is updated to \(A(t)\).
When the accuracy does not improve for $\gamma_{\text{thr}}$ consecutive rounds (i.e., $\tau_{p} = \gamma_{\text{thr}}$), the current round is regarded as the prestopping point, denoted as $T_e$.
At the end of Phase 1, the centralized server instructs the clients to apply loss correction for every round after $T_e$.
Empirical verification of the estimated accuracy, aggregated from each client's training performance, is provided in the left panel of \cref{fig:fig1}. 
In this panel, the estimated accuracy exhibits a sudden decline, attributed to the presence of noisy labels.

\subsection{Applying Loss Correction}
In Phase 2, clients involved in rounds after $T_e$ apply loss correction using enhanced forward correction. 
Before applying loss correction, clients generate a noise transition matrix based on their own datasets.
Each element of a noise transition matrix represents the probability of the true label given the observed label.
The columns of this matrix correspond to the true labels, while the rows represent the observed labels.
True labels are inferred with high confidence by the learning model whereas observed labels, which are potential noisy labels, are annotated in the dataset beforehand.
Each client applies loss correction using the noise transition matrix. 
This loss correction allows the loss function to operate as if the dataset were noise-free.

\subsubsection{Generating the Noise Transition Matrix}
The clients receiving the Phase 1 completion signal generate the noise transition matrix.
The noise transition matrix is derived from the count matrix, which consists of the number of examples for true and observed labels, as proposed in \cite{Curtis_2021_JAIR}.
The key difference between these matrices lies in whether they represent conditional probability or label quantities.
The count matrix is constructed with high reliability in a client-wise manner.
The main challenge of \texttt{FedEFC} is that it requires a pretrained model.
Without a pretrained model, the count matrix generated by the model currently being trained may consist of improper elements.

\begin{figure}[t] 
  \centering
  \includegraphics{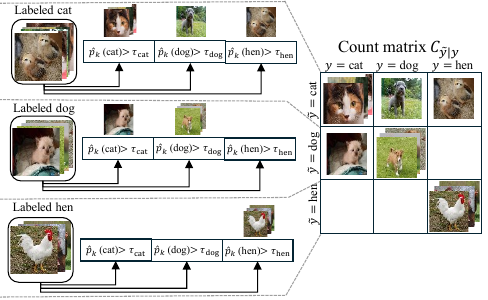}
  \caption{
  Example of generating the count matrix $C_{\tilde{y}|y}$. 
  The figure illustrates the process for three labels—cat, dog, and hen—where “labeled” indicates data annotated with the observed label, and the true label is determined based on the threshold $\tau_\text{label}$.
  }
  \label{fig:fig2}
\end{figure}

However, in the FL setting, utilizing the global model trained up to the prestopping point helps improve the alignment between the estimated and real noise matrices, as demonstrated in the right panel of \cref{fig:fig1}.
At the prestopping point, the cosine similarity confirms that the noise matrices predicted by the pretrained model and the FL global model are nearly identical to the real noise matrix.
Thus, the noise transition matrix is constructed by applying the global model after the prestopping point. 

Each client acquires the noise transition matrix as follows.
For client \(k\), the global model after round \(t = T_e\) predicts the label for input $\boldsymbol{x}^k$ as
\begin{align}
j = \arg\max_{j'\in\mathcal{C}} \hat{p}_{k}(\tilde{y}^{k}=j'|\boldsymbol{x}^k,w^t_g),
\end{align}
where \(\hat{p}_k(\cdot)\) denotes an estimate of the conditional probability for client \(k\).
We define the set of examples with true label \(y^k=j\) and observed label \(\tilde{y}^k=i\) as
\begin{align}
&\mathbf{X}_{\tilde{y}^k=i, y^k=j} \notag \\
 &\quad \quad= \Bigl\{ \boldsymbol{x}^k \in \mathbf{X}_{\tilde{y}^k=i} : \hat{p}_k(\tilde{y}^k=j | \boldsymbol{x}^k; w^{t}_g) \geq \tau_{j} \Bigr\},
\end{align}
where the set \(\mathbf{X}_{\tilde{y}^k=i}\) represents all data samples with observed label \(i\) in the dataset \(\mathcal{D}_k\) of client \(k\). 
The example set $\mathbf{X}_{\tilde{y}^k=i} $ satisfies $\mathbf{X}_{\tilde{y}^k=i, y^k=j} \subseteq \mathbf{X}_{\tilde{y}^k=i} \subseteq \mathcal{D}_k$.
The threshold \(\tau_{j}\) serves as the criterion for counting examples and is defined as
\begin{equation}
\tau_{j} = \frac{1}{\left| \mathbf{X}_{\tilde{y}^k=j} \right|} \sum_{\boldsymbol{x}^k \in \mathbf{X}_{\tilde{y}^k=j}} \hat{p}_k(\tilde{y}^k=j | \boldsymbol{x}^k; w^{t}_g),
\end{equation}
Threshold \(\tau_{j}\) is the mean estimation probabilities for data labeled as \(j\).
Using the example set \(\mathbf{X}_{\tilde{y}^k=i, y^k=j}\), the count matrix for client \(k\) is defined as $\boldsymbol{C}_{\tilde{y}^k=i, y^k=j} = \left| \mathbf{X}_{\tilde{y}^k=i, y^k=j} \right|$.
For simplicity, the notation is rewritten as
$\boldsymbol{C}_{\tilde{y}^k=i, y^k=j} \coloneqq \boldsymbol{C}^{k}_{\tilde{y}=i, y=j},
\mathbf{X}_{\tilde{y}^{k}=i, y^{k}=j} \coloneqq \mathbf{X}^k_{\tilde{y}=i, y=j}.$
The count matrix is constructed to enumerate the number of examples associated with each label pair, with columns representing true labels and rows indicating observed labels.
\cref{fig:fig2} illustrates an example of count matrix construction.
The noise transition matrix of client $k$ is computed as
\begin{equation}\label{eq:jointDistribution}
\boldsymbol{Q}^{k}_{\tilde{y}=i| y=j} = \frac{\boldsymbol{C}^{k}_{\tilde{y}=i, y=j}}{\sum_{j\in \mathcal{C}} \boldsymbol{C}^{k}_{\tilde{y}=i, y=j}}.
\end{equation}
\(\boldsymbol{Q}^{k}_{\tilde{y}=i| y=j}\) represents the \((i,j)\)th entry of the noise transition matrix \(\boldsymbol{Q}^k_{{\tilde{y}}|{y}}\). 
This element corresponds to an estimate of the conditional probability \(p(\tilde{y}^k=i| y^k=j)\), which indicates the probability that the observed noisy label \(\tilde{y}\) is \(i\) given that the true label \(y\) is \(j\).
Here, we assume that label noise is sample-independent, so \(p(\tilde{y}^k=i| y^k=j,\boldsymbol{x}^k) = p(\tilde{y}^k=i| y^k=j)\).
Consequently, each client generates its own noise transition matrix \(\boldsymbol{Q}^k_{{\tilde{y}}|{y}}\).

\subsubsection{Applying Loss Correction}
Participating clients apply loss correction based on the derived matrix $\boldsymbol{Q}_{{\tilde{y}}|{y}}$, following a similar approach to \cite{Patrini_2017_CVPR}, where loss functions are modified.
Regarding client $k$ in round $t$, the per-sample loss function $\mathcal{L}_k(\cdot)$ is given by $\mathcal{L}_k(\boldsymbol{x}, y; {w}_k^t)$.
When a client trains with a DNN and applies the softmax function \(\sigma_k(\cdot)\) in the final layer, the estimated conditional probability that the observed label is $\tilde{y}^{k}=i$ given input $\boldsymbol{x}^k$, using the client's local model weights $w=w_k^t$ as our estimator parameterization, is
\begin{align}
\hat{p}_k(\tilde{y} = i|\boldsymbol{x}^k; w_k^t) &= {\sigma_k}(\boldsymbol{h}_i^k(\boldsymbol{x}^k;w_k^t)) \\
&= \frac{{\exp}({\boldsymbol{h}_i^k(\boldsymbol{x};w_k^t)})}{\sum_{j \in \mathcal{C}} \exp({\boldsymbol{h}_{j}^k(\boldsymbol{x};w_k^t)})}.
\end{align}
\(\boldsymbol{h}_i^k \) denotes the output of the DNN at the final layer corresponding to label \( i \). 
The per-sample loss function \(\mathcal{L}_{k}\) can be rewritten using cross-entropy as
\begin{equation}\label{eq:crossentropy}
\mathcal{L}_k(\boldsymbol{x}^k, i; w_k^t) = -\log\hat{p}_k(\tilde{y}=i|\boldsymbol{x}^k; w_k^t).
\end{equation}
To streamline notation, we replace \(\boldsymbol{x}^k\), \(y^k\) and \(\tilde{y}^k\) with \(\boldsymbol{x}\), \(y\) and \(\tilde{y}\), respectively.
Let $\hat{p}_k(y=j|\boldsymbol{x}; w)$ be an estimator for the conditional probability 
that the true label is $y=j$, given input $\boldsymbol{x}$, where this estimator is parametrized by $w$. 
Since we assumed $p(\tilde{y}|y, \boldsymbol{x}) = p(\tilde{y}| y)$, it follows from Bayes' rule that
$p(\tilde{y}=i,y=j|\boldsymbol{x}) = p(\tilde{y}=i|y=j) \, p(y=j|\boldsymbol{x})$ which implies 
\begin{align}
  p(\tilde{y}=i | \boldsymbol{x}) = \sum_{j \in \mathcal{C}} p(\tilde{y}=i | y=j) \, p(y=j | \boldsymbol{x}).
\end{align}
Thus, using clients local model weights $w=w_k^t$ as our estimator parametrization, and using the noise transition matrix entry $Q^{k}_{\tilde{y}=i|y=j}$ as our estimate for $p(\tilde{y}=i|y=j)$, we shall define the estimator
\begin{align}
\!\!\!\!\hat{p}_k({y}=i | \boldsymbol{x}; w_k^t) := \sum_{j \in \mathcal{C}} Q^k_{\tilde{y}=i | y=j} \cdot \hat{p}_k({y}=j | \boldsymbol{x}; w_k^t),
\end{align}
and use this estimator to introduce an updated per-sample loss function $\mathcal{L}^{\textbf{F}}_k(\cdot)$, defined by
\begin{align}\label{eq:modifiedLossfcn}
  \mathcal{L}^{\textbf{F}}_k(\boldsymbol{x}^k, i; w_k^t) := - \log \hat{p}_k(y=i | \boldsymbol{x}; w_k^t).
\end{align}
Clients then begin transmitting their learning parameters, updated via loss correction, to the centralized server. 
The entire process is summarized in Algorithm \ref{alg:FedModified}.

\begin{algorithm}[t]
    \caption{FedEFC: Federated Learning Using Enhanced Forward Correction}
    \label{alg:FedModified}
    \begin{algorithmic}[1]
        \Require Global model $w_g^{0}$, maximum global epochs $T$
        \Ensure Final global model $w_{g}^T$

        \State {Phase 1: Find prestopping point}
        \State Initialize parameters $\tau_p = 0$ and $A_\text{max} = 0$
        \For{$t = 0$ to $T$}
            \State $\mathcal{S}_t \gets \text{Sample a subset of clients from } \{1,2,\dots,N\}$
            \For{each client $k \in \mathcal{S}_t$ in parallel}
                \State $\left(w_{k}^{t+1}, A_k(w_g^t)\right) \gets \text{Client update}(k, w_g^{t})$
                \State Compute $A(t)$ via Eq.~\eqref{eq:accuracy}
                \If{$A(t) > A_\text{max}$}
                    \State $\tau_p \gets 0$, $A_\text{max} \gets A(t)$
                \Else
                    \State $\tau_p \gets \tau_p + 1$
                    \If{$\tau_p = \gamma_{\text{thr}}$}
                        \State $T_e \gets t$
                        \State \textbf{break}
                    \EndIf
                \EndIf
            \EndFor
            \State $w_g^{t+1} \gets \sum_{k \in \mathcal{S}_t} \frac{n_k}{\sum_{u \in \mathcal{S}_t} n_u} w_{k}^{t+1}$
        \EndFor

        \State {Phase 2: Apply loss correction}
        \For{$t = T_e$ to $T$}
            \State $\mathcal{S}_t \gets \text{Sample a subset of clients from } \{1,2,\dots,N\}$
            \For{each client $k \in \mathcal{S}_t$ in parallel}
                \State Compute ${Q}^{k}_{y=j|\tilde{y}=i}$ via Eq.~\eqref{eq:jointDistribution}
                \State Update $\mathcal{L}_{k}^{\textbf{F}}(\boldsymbol{x}, y; w_k^t)$ via Eq.~\eqref{eq:modifiedLossfcn}
                \State $w_{k}^{t+1} \gets \text{Client update}(k, w_g^{t}, \mathcal{L}_{k}^{\textbf{F}}(\cdot))$
            \EndFor
            \State $w_g^{t} \gets \sum_{k \in \mathcal{S}_t} \frac{n_k}{\sum_{u \in \mathcal{S}_t} n_u} w_{k}^{t}$
        \EndFor
    \end{algorithmic}
\end{algorithm}

\subsection{Objective Function Analysis in Noisy Labels}
We analyze the objective function of FL in noisy labels when applying \texttt{FedEFC}.
Each participating client updates its model by minimizing the local loss function during each round.
Clients optimize their models iteratively. 
When using stochastic gradient descent (SGD), the update rule is given by ${w}_k^{t+1} \leftarrow {w}_k^t-\eta_k\nabla \mathcal{L}_k(\boldsymbol{x}, y; {w}_k^t)$ where $\eta_k$ denotes the learning rate of client $k$.
According to \cite{mcmahan2017communication}, the objective of FL is formulated as the minimization problem: Compute the value of
\begin{align}
\arg \min_{w} &{F}_g({w}) = \sum_{k\in \mathcal{S}_t} \frac{n_k}{\sum_{u\in\mathcal{S}_t}n_u}\mathcal{L}_k(\boldsymbol{x}, y; w).
\end{align}

The objective function in FL with noisy labels can be formulated using a composite loss, which combines a proper loss with a link function.
A proper loss is used for class probability estimation, while a link function maps the classifier's output to the range $[0,1]$.
A composite loss is a proper loss, as proven in \cite{Reid_2010_JMLR}.
In the FL system, local loss function corresponds to a proper loss and the inverse of the softmax function acts as the link function ($\sigma_{k}^{-1}:[0,1]\rightarrow\mathbb{R}$).
Since composite loss is defined as the combination of a proper loss and a link function, the composite loss $(\mathcal{L}_{k})^{\sigma^{-1}_k}$ for client $k$ is defined by
$(\mathcal{L}_{k})^{\sigma^{-1}_k}({y},{\boldsymbol{h}^{k}(\boldsymbol{x}};w)) = \mathcal{L}_{k}(\boldsymbol{x},y;w)$.
In the case of a clean dataset, the minimizer of the loss function, following the property of composite proper losses, is given by
\begin{align}
\!\!\!\!\!\!\arg\min_{\boldsymbol{h}} \mathbb{E}_{\boldsymbol{x},{y}}[(\mathcal{L}_{k})^{\sigma^{-1}_k}({y},\boldsymbol{h}(\boldsymbol{x}; w))]
= \sigma_k^{-1}(p({y}|\boldsymbol{x}; w)).
\end{align}
Therefore, the objective of FL can be reformulated as the following minimization problem: Compute the value of
\begin{align}
\arg \min_{w} \Bigg( \sum_{k\in \mathcal{S}_t} \frac{n_k\arg\min_{\boldsymbol{h}} \mathbb{E}_{\boldsymbol{x},{y}}[(\mathcal{L}_{k})^{\sigma^{-1}_k}({y},\boldsymbol{h}(\boldsymbol{x}; w))]}{\sum_{u\in\mathcal{S}_t}n_u}\Bigg).
\end{align}
For a noisy dataset, the minimizer of the composite loss $(\mathcal{L}_{k})^{\sigma^{-1}_k}$ is instead given by $\sigma_k^{-1}(p(\tilde{y}|\boldsymbol{x}; w))$.
However, under \texttt{FedEFC}, the objective evaluated at the minimizer for clean data remains approximately equal to that for noisy data when enhanced forward correction is applied.
\textbf{Thm.~\ref{thm:proper_composite}} shows this approximate equivalence, as derived in \cite{Patrini_2017_CVPR}.

\begin{theorem} \label{thm:proper_composite}
Assume each matrix \( Q^k_{\tilde{y}|y} \) generated by client $k$ is non-singular, and approximately equal to the true transition matrix $T^k_{\tilde{y}|y}$ whose $(i,j)$th entries are true conditional probabilities $p(\tilde{y}=i|y=j)$.
A composite loss incorporating \( Q^{k}_{{\tilde{y}}|{y}} \) is given by
\begin{align}
(\mathcal{L}_{k}^{\mathbf F})^{\sigma^{-1}_k}(y, \boldsymbol{h}^{k}(\boldsymbol{x};w))=\mathcal{L}_{k}^{\mathbf F}(\boldsymbol{x},y;w).
\end{align}
Then, the aggregated objective at the minimizer for clean data is approximately equal to that
at the minimizer for noisy data:
\begin{align}
&\sum_{k \in \mathcal{S}_t} \frac{n_k\,\arg\min_{\boldsymbol{h}}\mathbb{E}_{\boldsymbol{x},\tilde{y}}
   \bigl[(\mathcal{L}_{k}^{\mathbf F})^{\sigma^{-1}_k}(y,\boldsymbol{h}(\boldsymbol{x}; w))\bigr]}
   {\sum_{u \in \mathcal{S}_t}n_u}\\ \notag
&\;\;\approx\; \sum_{k \in \mathcal{S}_t} \frac{n_k\,\arg\min_{\boldsymbol{h}}\mathbb{E}_{\boldsymbol{x},y}
   \bigl[(\mathcal{L}_{k})^{\sigma^{-1}_k}(y,\boldsymbol{h}(\boldsymbol{x}; w))\bigr]}
   {\sum_{u \in \mathcal{S}_t}n_u}.
\end{align}
\end{theorem}

\noindent\textit{Proof.}
Fix client $k$ and define $\phi_k=Q^k_{{\tilde{y}}|{y}}\cdot\sigma_k$.
Then
\begin{align}
&\arg\min_{\boldsymbol{h}} \mathbb{E}_{\boldsymbol{x},\tilde{y}}
   \bigl[(\mathcal{L}_{k}^{\mathbf F})^{\sigma^{-1}_k}(y,\boldsymbol{h}(\boldsymbol{x}; w))\bigr]\\
&=\arg\min_{\boldsymbol{h}} \mathbb{E}_{\boldsymbol{x},\tilde{y}}
   \bigl[(\mathcal{L}_{k})^{\phi^{-1}_k}(y,\boldsymbol{h}(\boldsymbol{x}; w))\bigr]\\
&=\phi_k^{-1}(p(\tilde y|\boldsymbol{x})) =\sigma_k^{-1}\!\left((Q^{k}_{{\tilde{y}}|{y}})^{-1}p(\tilde y|\boldsymbol{x}) \right)\\
&=\sigma_k^{-1}\!\left((Q^{k}_{{\tilde{y}}|{y}})^{-1}T^{k}_{{\tilde{y}}|{y}}\,p(y|\boldsymbol{x}) \right) \approx \sigma_k^{-1}\!\left(I\cdot p(y|\boldsymbol{x})\right) \\
&=\arg\min_{\boldsymbol{h}} \mathbb{E}_{\boldsymbol{x},y}
   \bigl[(\mathcal{L}_{k})^{\sigma^{-1}_k}(y,\boldsymbol{h}(\boldsymbol{x}; w))\bigr].
\end{align}
For client $k$, the minimizer of the loss function using enhanced forward correction under noisy labels is shown to be approximately equivalent to the minimizer of the loss function under clean labels.
Since the summation of the minimizers forms the objective function, \textbf{Thm.~\ref{thm:proper_composite}} holds. 

\noindent\textbf{Remark.}
The accuracy of the approximation depends on how close $Q^k_{{\tilde{y}}|{y}}$ is to $T^k_{{\tilde{y}}|{y}}$.

\section{Experiments}\label{sec:experiments}
In this section, \texttt{FedEFC} is evaluated on non-IID data with noisy labels using the MNIST \cite{lecun1998gradient}, CIFAR-10 \cite{krizhevsky2009learning}, and CIFAR-100 \cite{netzer2011reading} datasets. 
To assess robustness against noisy labels, we vary noise parameters $(\rho, \zeta)$ with fixed heterogeneity and, conversely, vary heterogeneity parameters $(\alpha_{\text{dir}}, p)$ under fixed noise levels. 
This design enables simulation of environments from nearly IID to highly non-IID.
By combining these parameters, diverse scenarios with varying  $(\alpha_{\text{dir}},p)$ and $(\rho, \zeta)$ are constructed. 
We compare \texttt{FedEFC} with standard FL techniques, noise-robust FL approach, and CL-based methods originally developed for DNNs trained with noisy labels but extended to the FL setting.
The results demonstrate that \texttt{FedEFC} is robust against both noisy labels and data heterogeneity.

\subsection{Experiment Setup}
\textbf{Non-IID with Noisy Labels}:
To emulate a real-world scenario, noise is first introduced into the training dataset before it is allocated to all clients in the FL system. 
Since similar noisy labels occur across clients, a consistent noise pattern is applied before partitioning the dataset according to non-IID parameters $(\alpha_\text{dir},p)$.
This procedure creates a non-IID environment in which noisy labels are class-dependent.

\textbf{Baselines}:
We compare \texttt{FedEFC} with standard FL methods (\texttt{FedAvg}~\cite{mcmahan2017communication}, 
\texttt{FedDitto}~\cite{li2021ditto}, \texttt{FedProx}~\cite{li2020fedprox}), noise-robust baselines (\texttt{FedCorr}~\cite{Xu_2022_CVPR}, \texttt{FedNoRo}~\cite{wu2023fednoronoiserobustfederatedlearning}, 
\texttt{RoFL}~\cite{yang2022robust}), and, once the prestopping point is determined, loss correction (\texttt{forward correction}~\cite{Patrini_2017_CVPR}) and data pruning (\texttt{confident learning}~\cite{Curtis_2021_JAIR}) methods adapted to FL.  
As a reference, \texttt{FedAvg} without noise is reported as the noise-free upper bound.

\subsection{Implementation Details}\label{ex:detail}
FL experiments are simulated with $N = 100$ clients. 
Each client performs local training using SGD with a momentum of $0.5$ and at each round, a fraction $0.1$ of clients are randomly selected for aggregation.
For MNIST, we use a 9-layer CNN trained for $100$ rounds with batch size $64$; for CIFAR-10, ResNet-18 for $200$ rounds with batch size $10$; and for CIFAR-100, ResNet-34 for $300$ rounds with batch size $10$.
The prestopping threshold, $\gamma_\text{thr}$, is empirically set to $3$ for MNIST and $6$ for CIFAR-10 and CIFAR-100. 
To ensure training stability, prestopping monitoring begins after round $10$ for MNIST and after round $40$ for CIFAR-10 and CIFAR-100.

In cases of severe label imbalance, particularly when $p=0.2$, estimating the noise transition matrix directly from the raw count matrix is challenging. 
To compensate for this issue, we apply a weighted noise transition matrix defined as $\widetilde{Q}^k_{\tilde{y}=i \mid y=j} = \frac{p^k_i\,\boldsymbol{C}^k_{\tilde{y}=i,y=j}}{\sum_{j} p^k_j\,\boldsymbol{C}^k_{\tilde{y}=i,y=j}}$ where $p^k_i = \frac{|\mathbf{X}_{\tilde{y}^k=i}|}{\sum_j |\mathbf{X}_{\tilde{y}^k=j}|}$. 
This weighting scheme compensates for label imbalance by incorporating each client's class distribution.
The adjustment is applied to MNIST and CIFAR-10 when $p=0.2$; otherwise, the standard count matrix remains effective.

\begin{table*}
  \centering
  \caption{Average test accuracy ($\pm$ std over 3 trials) under noise-fixed settings with $\rho=0.2$, $\zeta=0.8$, across different non-IID parameters. The best two values per column are in bold, except for the noiseless case (FedAvg wo noise).}
  \label{tab:noise_fixed_summary}
  \resizebox{\textwidth}{!}{%
    \begin{tabular}{l|ccc|ccc|ccc}
      \toprule
        Dataset & \multicolumn{3}{c|}{MNIST} & \multicolumn{3}{c|}{CIFAR-10} & \multicolumn{3}{c}{CIFAR-100} \\
        (\(\alpha_{\text{dir}}\), \(p\)) & (100.0, 0.8) & (10.0, 0.5) & (1.0, 0.2) & (100.0, 0.8) & (10.0, 0.5) & (1.0, 0.2) & (100.0, 0.8) & (10.0, 0.5) & (1.0, 0.2) \\
      \midrule
      FedAvg      & \(97.07\pm0.10\)  & \(96.31\pm0.42\)  & \(84.13\pm2.23\)  & \(82.57\pm0.55\)  & \(78.49\pm0.75\)  & \(61.27\pm2.81\)  &  $55.30\pm0.23$ & $54.97\pm0.31$   & $46.09\pm0.35$  \\
      FedDitto    & \(97.00\pm0.05\)  & \(96.44\pm0.39\)  & \(85.26\pm3.61\)  & \(83.53\pm0.15\)  & \(80.48\pm0.95\)  & \(59.53\pm3.78\)  & $53.59\pm0.37$     & $53.91\pm0.26$ & $45.27\pm0.15$ \\
      FedProx     & \(97.11\pm0.12\)  & \({96.25\pm0.58}\)  & \(84.76\pm2.34\)  & \(82.34\pm0.57\)  & \(78.77\pm0.71\)  & \(61.50\pm2.16\)   & $55.68\pm0.71$& $54.24\pm0.43$    & $44.73\pm0.69$  \\
      \midrule
      FedCorr     & \(96.47\pm0.64\)  & \(93.61\pm1.16\)  & \(90.69\pm0.73\)  & \({84.90\pm0.08}\)  & \(72.54\pm4.86\)  & \(63.71\pm0.73\) & $\bm{60.11\pm0.98}$& $\bm{57.31\pm0.31}$   & $44.80\pm0.42$ \\      RoFL&$\bm{99.00\pm0.05}$&$\bm{98.98\pm0.05}$&$\bm{97.36\pm0.50}$&$\bm{89.22\pm0.16}$&$\bm{86.11\pm0.13}$&$54.65\pm0.31$& $58.55\pm0.04$&$51.33\pm0.45$&$40.97\pm0.85$\\ 
      FedNoRo &$97.43\pm0.13$&$95.56\pm1.15$&$65.08\pm4.62$&$84.19\pm0.25$& $76.68\pm2.44$ &$37.54\pm2.30$&${58.57\pm1.27}$&$43.07\pm0.64$&$14.57\pm1.94$\\
      \midrule
      confident learning
                  & $\bm{98.97\pm0.13}$  & $\bm{98.94\pm0.16}$  &  $91.44\pm2.75$ & ${88.01\pm0.24}$ & $80.21\pm3.29$  &  $62.45\pm1.21$ & ${57.92\pm0.11}$ & $56.68\pm0.83$ & $47.49\pm1.08$\\
      forward correction 
                  & ${97.05\pm0.52}$  & \(94.79\pm0.09\)  & \(\bm{97.00\pm0.20}\)  & \(82.76\pm0.21\)  & \({80.62\pm0.70}\)  & \(\bm{70.80\pm1.59}\)  & \ $56.72\pm0.58$   & $56.73\pm0.22$  & $\bm{52.00\pm0.94}$ \\
      FedEFC 
                  & ${98.15\pm0.45}$ & ${98.51\pm0.46}$ & ${96.78\pm0.60}$ & $\bm{88.37\pm0.08}$ & $\bm{86.78\pm0.36}$ & $\bm{70.57\pm1.25}$ & $\bm{60.57\pm0.40}$ & $\bm{59.34\pm0.51}$ & $\bm{50.58\pm0.18}$\\
     \midrule
      FedAvg wo noise
                  & \(99.49\pm0.01\)  & \(99.51\pm0.02\)  & \(98.91\pm0.21\)  & \(90.74\pm0.14\)  & \(88.93\pm0.46\)  & \(80.43\pm0.86\)  &  $68.14\pm0.51$& $67.42\pm0.32$& $63.58\pm0.20$ \\
      \bottomrule
    \end{tabular}%
  }
\end{table*}

\begin{table*}
  \centering
  \caption{Average test accuracy ($\pm$ std over 3 trials) under non-IID fixed settings 
$(\alpha_{\text{dir}}, p) = (10.0, 0.5)$ with varying noise configurations.}
  \label{tab:heterogeneous_fixed_summary}
  \resizebox{\textwidth}{!}{%
    \begin{tabular}{l|ccc|ccc|ccc}
      \toprule
        Dataset & \multicolumn{3}{c|}{MNIST} & \multicolumn{3}{c|}{CIFAR-10} & \multicolumn{3}{c}{CIFAR-100} \\
        (\(\rho\), \(\zeta\)) & (0.4, 0.8) & (0.2, 0.4) & (0.1, 0.0) & (0.4, 0.8) & (0.2, 0.4) & (0.1, 0.0) & (0.4, 0.8) & (0.2, 0.4) & (0.1, 0.0) \\
      \midrule
      FedAvg      
                  & \(66.63\pm0.26\)  & \(96.94\pm0.12\)  & \(98.74\pm0.02\)  
                  & \(58.42\pm0.24\)  & \(79.22\pm0.29\)  & \(84.83\pm0.21\)  
                  & \(38.04\pm0.59\)  & \(55.45\pm1.06\)  & \(64.55\pm0.49\) \\
      FedDitto    
                  & \(66.70\pm0.31\)  & \(97.09\pm0.16\)  & \(98.64\pm0.04\)  
                  & \(59.18\pm0.44\)  & \(80.11\pm0.72\)  & \(83.09\pm0.09\)  
                  & \(38.02\pm0.40\)  & \(54.26\pm0.61\)  & \(61.95\pm0.47\) \\
      FedProx     
                  & \(67.13\pm0.31\)  & \(97.03\pm0.12\)  & \(98.82\pm0.09\)  
                  & \(58.78\pm0.14\)  & \(79.24\pm0.31\)  & \(85.03\pm0.06\)  
                  & \(38.18\pm0.40\)  & \(55.68\pm0.67\)  & \(64.82\pm0.22\) \\
      \midrule
      FedCorr     
                  & \(65.71\pm0.49\)  & \(97.36\pm0.60\)  & \(\bm{99.19\pm0.13}\)  
                  & \(54.06\pm27.01\) & \(78.10\pm1.42\)  & \(80.31\pm2.65\)  
                  & \(\bm{44.86\pm0.07}\)  & \(\bm{58.13\pm1.57}\)  & \(64.50\pm1.78\) \\
    RoFL & $68.50\pm0.27$ &$\bm{99.21\pm0.05}$ &${99.14\pm0.01}$
        &$62.29\pm2.53$ & $\bm{86.91 \pm 0.42}$  &$\bm{87.85\pm0.14}$ 
        &$34.94\pm0.16$ & $52.83\pm0.21$ &$55.46\pm0.05$\\
        FedNoRo &$\bm{70.17\pm1.88}$&$97.23\pm0.26$&$98.73\pm0.06$&$58.94\pm2.15$&$78.59\pm0.70$&$83.47\pm0.15$&$37.63\pm0.64$&$43.38\pm2.50$&$50.63\pm1.43$\\    
        \midrule
      confident learning
                  & ${68.38\pm1.75}$ & $\bm{99.22\pm0.04}$  &  $\bm{99.33\pm0.01}$ & $\bm{66.20\pm1.40}$ & ${84.95\pm0.22}$ & ${86.83\pm0.26}$ & $41.37\pm1.26$& $56.19\pm0.62$& $64.51\pm0.24$\\
      
      forward correction
                  & \({67.69\pm0.84}\)  & \({97.45\pm0.28}\)  & \(98.84\pm0.08\)  
                  & \({60.61\pm1.47}\)  & \({80.76\pm0.59}\)  & \({85.38\pm0.11}\)  
                  & \(41.30\pm0.27\)  & $57.59\pm0.51$               & $\bm{65.02\pm0.07}$ \\
      FedEFC 
                  & \(\bm{71.35\pm3.68}\) & \({98.39\pm0.33}\) & \({99.04\pm0.04}\)  & \(\bm{85.85\pm0.66}\) & \(\bm{85.79\pm0.09}\) & \(\bm{87.12\pm0.16}\)  & \(\bm{47.17\pm0.59}\) & \(\bm{59.58\pm0.41}\) & $\bm{65.80\pm0.33}$\\
      \midrule
      FedAvg wo noise
                  & \multicolumn{3}{c|}{${99.51\pm0.02}$}  
                  &  \multicolumn{3}{c|}{${88.93\pm0.46}$}  
                  &  \multicolumn{3}{c}{${67.42\pm0.32}$} \\
      \bottomrule
    \end{tabular}%
  }
\end{table*}

\begin{table}
  \centering
  \caption{CIFAR-10 test accuracy ($\pm$ std, 3 runs) under high noise with $\rho=0.8$, $\zeta\in\{0.0,0.4,0.8\}$, and $(\alpha_{\text{dir}},p)=(100.0,0.8)$.}
  \label{tab:highnoise}
\resizebox{\linewidth}{!}{
    \begin{tabular}{l|ccc}
      \toprule
        (\(\rho, \zeta)\) & (0.8, 0.0) & (0.8, 0.4) & (0.8, 0.8)  \\
      \midrule
      FedAvg      & $21.88\pm0.48$ & $17.30\pm0.28$  & $18.75\pm0.66$ \\
      FedDitto    & $21.97\pm0.36$  & $14.28\pm0.58$  & $19.38\pm0.86$  \\
      FedProx     & $21.33\pm0.33$ & $17.14\pm0.10$  & $18.58\pm0.33$  \\
      \midrule
      FedCorr     & $20.54\pm0.67$  & $14.30\pm0.39$  & $9.54\pm1.31$  \\      
      RoFL        & $25.38\pm1.18$  & $15.51\pm0.21$ & $17.52\pm0.32$\\ 
      FedNoRo    &$\bm{25.92\pm1.82}$    & $\bm{20.31\pm0.04}$  & $\bm{27.70\pm1.22}$\\
      \midrule
      confident learning
                  & $19.25\pm0.53$ & $11.48\pm0.29$ & $18.49\pm0.12$\\
      forward correction 
                  & $23.40\pm0.31$ & $17.02\pm0.25$ & $18.61\pm0.24$\\
      FedEFC 
                  & $\bm{27.37\pm1.55}$ & $\bm{20.69\pm3.94}$&$\bm{29.90\pm3.06}$\\
      \bottomrule
    \end{tabular}}
\end{table}

\subsection{Comparison with Baselines}
\textbf{Non-IID variation}:
To evaluate the impact of non-IID conditions, we fix noise at $(\rho,\zeta)=(0.2,0.8)$ and vary $(\alpha_{\text{dir}}, p)$ across three configurations: $(100.0, 0.8)$, $(10.0, 0.5)$, and $(1.0, 0.2)$ (see Table~\ref{tab:noise_fixed_summary}).
\texttt{FedEFC} consistently achieves the best performance across most settings.
On MNIST, \texttt{RoFL} and \texttt{confident learning} achieve slightly higher accuracy under certain non-IID levels, but their gains over \texttt{FedEFC} are within 1\%.
\texttt{forward correction} performs best under the most severe case $(1.0, 0.2)$, yet its accuracy remains comparable to \texttt{FedEFC}, with gaps under 0.3\% on MNIST and CIFAR-10, and 1.5\% on CIFAR-100.
Overall, \texttt{FedEFC} maintains superiority, delivering stable and resilient performance in the non-IID FL condition.

\textbf{Noise variation}:
Noise robustness is examined by fixing data allocation and varying noise settings across $(\rho,\zeta)$ pairs: $(0.4,0.8)$, $(0.2,0.4)$, and $(0.1,0.0)$ (see Table~\ref{tab:heterogeneous_fixed_summary}).
In most noise conditions, \texttt{FedEFC} demonstrates greater robustness, steadily surpassing the baseline methods, with only minor differences in a few cases.
For the low-noise setting $(0.1,0.0)$ on MNIST, \texttt{FedCorr} and \texttt{confident learning} marginally outperform \texttt{FedEFC}, with differences of less than 0.3\%.  
Under the moderate setting $(0.2,0.4)$ on MNIST, \texttt{confident learning} achieves the highest accuracy, but remains within a 1\% margin of \texttt{FedEFC}.  
In the high-noise setting $(0.4,0.8)$ on CIFAR-10, \texttt{FedEFC} experiences only a 3\% performance drop compared to \texttt{FedAvg} in a noise-free setting.

\textbf{Extreme high noise}: 
To examine robustness under highly adverse conditions, we evaluate CIFAR-10 with extreme noise $(\rho,\zeta)\in\{(0.8,0.0),(0.8,0.4),(0.8,0.8)\}$ (see Table~\ref{tab:highnoise}).
While the standard deviation across trials is higher in this setting, \texttt{FedEFC} remains the most robust method, achieving the highest accuracy in all cases.
At $(0.8,0.4)$, the margin over the second-best method \texttt{FedNoRo} is minimal, remaining below $0.5\%$.
In contrast, the largest margin occurs at $(0.8,0.8)$, where \texttt{FedEFC} outperforms \texttt{FedNoRo} by more than $2\%$, while maintaining consistent gains across all extreme-noise cases.
Detailed configurations and additional results covering \textit{all cases} of $(\rho,\zeta)$ and $(\alpha_{\text{dir}}, p)$ are provided in the supplementary material.
\section{Discussion}\label{sec:discussion}
\textbf{Enhanced forward correction}: 
The key distinction between \texttt{forward correction} and \texttt{FedEFC} lies in how the noise transition matrix is constructed. 
\texttt{FedEFC} derives it by counting samples that satisfy a confidence threshold, whereas \texttt{forward correction} uses predicted probabilities (e.g., 97th percentile in \cite{Patrini_2017_CVPR}). 
Unlike probability-based estimation, the count-based approach is more resilient to prediction errors \cite{Curtis_2021_JAIR}. 
In FL with non-IID data, where some labels may be absent locally, we further adjust the count matrix using class distribution weights in \cref{ex:detail}. 

\textbf{Global model in FL}:
A key challenge in constructing a noise transition matrix is the conventional requirement for a pretrained model, since using a training model often leads to overfitting on the same data. 
For example, \cite{chen2019understanding} employs iterative noise cross-validation, partitioning the dataset to avoid this issue. 
In contrast, the global model in FL is not tied to any single client’s data, enabling effective construction of the noise transition matrix without a pretrained model or data partitioning. 
\texttt{FedEFC} leverages this property to address the challenge without additional procedures.

\textbf{Clients with heterogeneous noise levels}: 
Another practically relevant scenario arises when different clients are subject to different noise conditions. 
We evaluated CIFAR-10 under $(\alpha_{\text{dir}}, p)=(100.0,0.8)$, where half the clients followed $(\rho,\zeta)=(0.2,0.4)$ and $(\rho,\zeta)=(0.4,0.8)$. 
In this heterogeneous setup, \texttt{FedEFC} achieved the highest accuracy, demonstrating considerable robustness even when noise patterns varied across clients.
Detailed results for this case are provided in the supplementary material.

\section{Conclusion}\label{sec:conclusion}
We propose \texttt{FedEFC}, a novel algorithm to address the noisy label problem in FL systems with non-IID data distributions. 
It consists of two phases: (1) finding a prestopping point and (2) applying loss correction.
A key advantage of \texttt{FedEFC} is that it effectively mitigates the impact of noisy labels without requiring any inter-client information exchange. 
In particular, theoretical analysis establishes that, under \texttt{FedEFC}, the FL objective with noisy labels is equivalent to that with clean labels. 
Extensive experiments across varying noise levels and data heterogeneity demonstrates that \texttt{FedEFC} consistently outperforms conventional FL algorithms. 
Given that \texttt{FedEFC} operates under realistic conditions with noisy labels and non-IID data, it has significant potential for adoption across diverse FL scenarios.
\section*{Acknowledgments}\label{sec:Acknowledgments}
This research was partly supported by the Institute of Information \& Communications Technology Planning \& Evaluation (IITP)-ITRC (Information Technology Research Center) grant funded by the Korea government (MSIT) (IITP-2026-RS-2020-II201787, contribution rate: 50\%).
Also, this work was supported in part by the National Research Foundation of Korea (NRF) grant funded by the Korea government (MSIT) (No. RS-2023-00212836, contribution rate: 50\%).
{
    \small
    \bibliographystyle{ieeenat_fullname}
    \bibliography{main}
}

\end{document}